%% file: sample-sigconf.tex
\documentclass[sigconf]{acmart}
\newcommand{\ignore}[1]{}
\usepackage{xspace}

\newcommand{\be}{\mathbf{e}}
\newcommand{\bm}{\mathbf{m}}

\newcommand{\bb}{\mathbf{b}}

\newcommand{\bW}{\mathbf{W}}

\usepackage{amsfonts,amssymb}
\usepackage{booktabs} 
\usepackage{subfigure}
\usepackage{caption}
\usepackage{bm}
\usepackage{multirow}
\usepackage{footmisc}
\usepackage{amssymb}
\usepackage{array}
\usepackage{graphicx}
\usepackage{clrscode}
\usepackage{balance}
\usepackage{float}
\usepackage{color}
\usepackage{xcolor}
\usepackage{amsopn}
\usepackage{mathrsfs}
\usepackage{mathtools}
\usepackage{amsmath}
\usepackage{booktabs}
\usepackage{arydshln}
\usepackage{hyperref}
\usepackage{blkarray}
\usepackage{enumerate}
\usepackage{courier}
\usepackage{mathrsfs}
\usepackage{rotating}
\usepackage{bm}
\usepackage{array}
\usepackage{ragged2e}
\usepackage{amssymb}
\usepackage{multirow}
\usepackage{multicol}
\usepackage{latexsym}
\usepackage{mflogo}
\usepackage{amsthm}
\usepackage[ruled,linesnumbered]{algorithm2e}

\newcommand{\paratitle}[1]{\vspace{1.5ex}\noindent\textbf{#1}}
\newcommand{\ie}{\emph{i.e.,}\xspace}

\newcommand{\eg}{\emph{e.g.,}\xspace}

\AtBeginDocument{%
  \providecommand\BibTeX{{%
    \normalfont B\kern-0.5em{\scshape i\kern-0.25em b}\kern-0.8em\TeX}}}

\copyrightyear{2022}
\acmYear{2022}
\setcopyright{acmcopyright}
\acmBooktitle{Proceedings of the Fifteenth ACM International Conference on Web
Search and Data Mining (WSDM '22), February 21--25, 2022, Tempe, AZ, USA}
\acmPrice{15.00}
\acmDOI{10.1145/3488560.3498514}
\acmISBN{978-1-4503-9132-0/22/02}

\acmConference[WSDM '22]{Proceedings of the Fifteenth
ACM International Conference on Web Search and Data Mining}{February 21--25,
2022}{Tempe, AZ, USA}

\begin{document}

\title{C$^2$-CRS: Coarse-to-Fine Contrastive Learning for\\
Conversational Recommender System}

\author{Yuanhang Zhou$^{1\dagger}$, Kun Zhou$^{1\dagger}$, Wayne Xin Zhao$^{2,3^*}$, Cheng Wang$^{4}$, Peng Jiang$^{4}$, He Hu$^{1}$}\thanks{$^\dagger$Equal contribution} \thanks{$^*$Corresponding author.}
\affiliation{%
 \institution{$^1$School of Information, Renmin University of China, China}
 \institution{$^2$Gaoling School of Artificial Intelligence, Renmin University of China, China}
 \institution{$^3$Beijing Key Laboratory of Big Data Management and Analysis Methods, China}
 \institution{$^4$Kuaishou Inc., China}
 \country{}
}
\affiliation{%
  \institution{sdzyh002@gmail.com, francis\_kun\_zhou@163.com, batmanfly@gmail.com, \\
  wangcheng03@kuaishou.com, jp2006@139.com, hehu@ruc.edu.cn}
 \country{}
}

\renewcommand{\shortauthors}{}

\fancyhead{} 

\begin{abstract}
Conversational recommender systems (CRS) aim to recommend suitable items to users through natural language conversations. For developing effective CRSs, a major technical issue is how to accurately infer user preference from very limited conversation context. To address issue, a promising solution is to incorporate external data for enriching the context information. However, prior studies mainly focus on designing fusion models tailored for some specific type of external data, which is not general to model and utilize multi-type external data.

To effectively leverage multi-type external data, we propose a novel coarse-to-fine contrastive learning framework to improve data semantic fusion for CRS. In our approach, we first extract and represent multi-grained semantic units from different data signals, and then align the associated multi-type semantic units in a coarse-to-fine way. To implement this framework, we design both coarse-grained and fine-grained procedures for modeling user preference, where the former focuses on more general, coarse-grained semantic fusion and the latter focuses on more specific, fine-grained semantic fusion. Such an approach can be extended to incorporate more kinds of external data. Extensive experiments on two public CRS datasets have demonstrated the effectiveness of our approach in both recommendation and conversation tasks.
\end{abstract}



\maketitle

\input{intro}
\input{related}
\input{problem}
\input{model}

\input{experiment}

\section{Conclusion and Future Work}
In this paper, we proposed a novel contrastive learning based coarse-to-fine pre-training approach for conversational recommender system. 
By utilizing the coarse-to-fine pre-training strategy, multi-type data representations can be effectively fused, such that the  representations for limited conversation context are further enhanced, which finally improve the performance of CRS.
By constructing extensive experiments, the effectiveness of our approach in both recommendation and conversation tasks has been demonstrated. It has shown that our approach is effective to bridge the semantic gap between different external data signals for CRS. 
Note that our approach is flexible to incorporate more kinds of external data, and is general to improve other tasks.

Currently, our focus is how to perform effective semantic fusion for incorporating external data for CRSs. As future work, we will consider designing a more general representation model that can be directly pretrained with various kinds of context data.

\begin{acks}
We are thankful to Xiaolei Wang for their supportive work and insightful suggestions. This work was partially supported by the National Natural Science Foundation of China under Grant No. 61872369 and 61832017, Beijing Academy of Artificial Intelligence (BAAI) under Grant No. BAAI2020ZJ0301 and Beijing Outstanding Young Scientist Program under Grant 
No. BJJWZYJH0120191000200\\98, and Public Computing Cloud, Renmin University of China. Xin Zhao is the corresponding author.
\end{acks}

\bibliographystyle{ACM-Reference-Format}
\balance
\bibliography{sample-base}

\end{document}

%% file: intro.tex
\section{Introduction}

\ignore{
\begin{table}
\caption{An illustrative example of a user-system conversation for movie recommendation.
The mentioned movies and important context words are marked in italic blue font and red font, respectively. }
\centering
\small
  \begin{tabular}{l|p{5.5cm}}
    \hline
    \texttt{User} & I am looking for some movies\\
    \texttt{System} &What kinds of movie do you like? \\
    \hline
    \texttt{User} & Today I'm in a mood for something \textcolor{red}{scary}. Any similar movies like \textcolor{blue}{\emph{Paranormal Activity (2007)}}? \\
    \hline
    \texttt{System} &\textcolor{blue}{\emph{It (2017)}} might be good  for you. It is a classic \textcolor{red}{thriller} movie with \textcolor{red}{good plot}. \\
    \hline
    \texttt{User} &Great! Thank you!\\
    \bottomrule
  \end{tabular}
  \label{case}
\label{example}
\end{table}}

\ignore{
Being able to provide personal suggestions, recommender systems~\cite{he2017neural,pazzani2007content,rendle2012bpr} have exhibited remarkable success in countless online applications.
They leverage user's historical behaviors (\eg rating history and click history) to predict user's preferred items.
Despite the success, it is still hard for recommender systems to capture user dynamic interests timely in online platform~\cite{hidasi2015session,kang2018self} or deal with newly coming users who have only limited historical behaviors~\cite{lika2014facing,schein2002methods}.
As an appealing solution, conversational recommender system (CRS)~\cite{SunZ18,Christakopoulou16,Li2018TowardsDC} has attracted widespread attention, which directly asks user preferences and provides user-specific suggestions or recommendation through natural language conversation.}

With the rapid development of intelligent agents in e-commerce platforms, conversational recommender system (CRS)~\cite{SunZ18,Christakopoulou16,Li2018TowardsDC} has become an emerging research topic, which aims to provide effective recommendations to users through natural language conversations.
In general, a CRS consists of a conversation module to converse with users and a recommender module to make recommendations.

To improve user satisfaction, a major objective of the CRS is to accomplish the recommendation task within
as few conversation turns as possible. 
Therefore, it is important for a CRS to accurately capture user preferences based on several initial turns of conversational utterances, which contain very limited context information to understand user needs.
Considering this issue, existing works~\cite{Chen2019TowardsKR,Liao2019DeepCR,ZhangCA0C18} have introduced external data sources to enrich the contextual information.
One line of research~\cite{Chen2019TowardsKR,zhou2020improving} utilizes  structured external data (\ie knowledge graphs) to enhance the representations of entities and words occurring in the conversational context.
Another line of research~\cite{lu2021revcore} introduces unstructured external data (\eg item reviews) to improve the item representations for recommendation and help generate informative responses.

\begin{figure}
\includegraphics[width=0.5\textwidth]{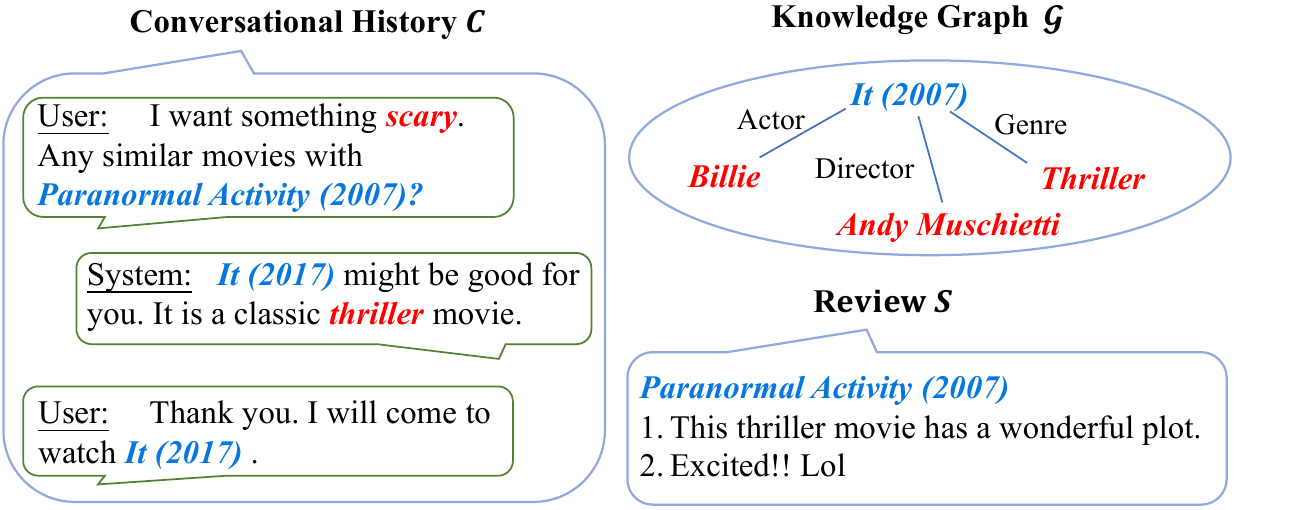}
\caption{An illustrative example of a conversation on movie recommendation between a user and the system. The associated external structured data (knowledge graph) and unstructured data (reviews) of conversational context have been also presented. Items (movies) are in blue and entities (\eg actors) are in red.}
\label{data}
\end{figure}


Due to the data heterogeneity, it is difficult to directly external data for improving CRS, because conversation data and external data usually correspond to very different information forms (\eg conversation utterances \emph{v.s.} knowledge graphs) or semantic content (\eg conversation utterance \emph{v.s.} online reviews). 
There exists a natural semantic gap between conversation data and external data.
The case becomes even more difficult when multi-type external data (also in different data forms) is available for utilization. Several efforts have been made to leverage external data for improving CRS~\cite{zhou2020improving,lu2021revcore}. However, they mainly focus on designing specific semantic fusion model tailored to some type of external data, which cannot apply to multi-type external data.  
Therefore, it is essential to develop a general approach to bridging the semantic gap between different data signals for CRS.



\ignore{Since these methods have shown the effectiveness of using external data, it is promising to combining the multi-type external data with the conversational context for improving the CRS.
Nevertheless, the following challenges make it hard to achieve this goal.
First, the multi-type external data is in different formats (\eg graph and text), which leads to a natural semantic gap to fuse the different information.
Although existing works have investigated to fuse unstructured data~\cite{zhou2020improving} or unstructured data~\cite{lu2021revcore}, it is still a hard problem to fuse the semantic information from structured and unstructured data.
Second, it is hard to capture the multi-grained user preference across the multi-type data.
In fact, a CRS requires to capture user's coarse-grained preference to select candidates from the large-scale item set, and the fine-grained preference to choose the most proper item for the recommendation. 
However, since the basic elements (\eg word or entity) of the multi-type data is various, user's coarse-grained and fine-grained preferences are usually represented differently across the different types of data.
It leads to another gap among the multi-type data that the multi-grained user preference are hard to capture and utilize for the CRS task.
Therefore, it is essential to consider how to bridge the semantic gap derived from the different data formats and effectively capture the multi-grained user preference within the multi-type data.}

To fuse multi-type context data, a major challenge is that they usually correspond to very different semantic spaces.  And, it may hurt the original representation performance if we directly align their semantic space as  previous studies~\cite{zhou2020improving,lu2021revcore}. 
To address this issue, we are inspired by an important observation that context data itself (no matter external data or conversation data) is in a multi-grained form, \eg \emph{entity} and \emph{entity subgraph} in knowledge graphs, and \emph{word} and \emph{sentence} in utterances.
Actually, user preference is also reflected in a multi-grained way~\cite{zhou2020s3}: a fine-grained semantic unit reflects some specific tastes (\eg an actor entity), while a coarse-grained semantic unit reflects some general tastes (\eg a set of comments for fiction movies). 
Given a conversation scenario, the semantic units from different data signals can be associated according to the reflected user preference. 
For example, in Figure~\ref{data}, the word ``\emph{It}'' in the fourth utterance corresponds to the entity ``\emph{It~(2017)}'' in  Freebase, and the fourth utterance essentially corresponds to a subgraph spanned based on the entity ``\emph{It~(2017)}''.
Such an example indicates that it needs a multi-grained semantic alignment in order to fuse different semantic spaces and better characterize user preference. 


\ignore{In fact, within the multi-type input data, the conversational context, unstructured and structured external data of the same user all reflect her/his overall preference from different views.
Therefore, it is promising to align the representations from the above views for bridging the semantic gap of the multi-type data.
Recently, contrastive learning has shown effectiveness in semantic fusion~\cite{} and representation alignment~\cite{}, which aims to learn unified and effective representation by pulling semantically close examples together and pushing apart irrelevant ones.
Fortunately, the representations of the same user from the different views are naturally semantic-close examples.
Contrastive learning can unify these representations by reducing their distance in the semantic space, so that the semantic gap derived from the different data formats can be alleviated.
To achieve it, one critical question in contrastive learning is how to construct positive and negative examples.
Consider that the multi-type data preserves multi-grained user preference, it seems that such a multi-grained information can be used to solve this problem: we can take the data representations with the same granularity level of the same user preference as the positive examples, and then utilize contrastive learning to pull them closer in the semantic space.
In this way, we can not only capture and utilizes the multi-grained user preference, but also unify and fuse the representations of the multi-type data.
Therefore, we propose a coarse-to-fine contrastive learning framework, where we first unify the representations of the coarse-grained information (\ie conversational context, unstructured and structured external data), and then align the representations of the fine-grained elements within the above data (\eg words and entites).
In this way, the data representations can incrementally learn the user preference from macro to micro perspectives, which encourages the multi-grained alignment and fusion of the multi-type input data.
}

To this end, in this paper, we propose a novel \textbf{C}oarse-to-fine \textbf{C}ontrastive learning approach for \textbf{C}onversational \textbf{R}ecommender \textbf{S}ystem, called \textbf{C$^2$-CRS}.
The core idea is to first extract and represent associated multi-grained semantic units from different data signals, and then align the corresponding semantic units from different data signals in a coarse-to-fine way.
To implement coarse-to-fine semantic fusion, we involve both \emph{coarse-grained} and \emph{fine-grained} pre-training procedures, where the former focuses on more general, coarse-grained user preference while the latter focuses on more specific, fine-grained user preference.
Such a way allows the model to gradually fuse different semantic spaces in a multi-grained manner, which is likely to yield more coherent fusion representations. 
The optimization objectives for coarse- and fine-grained fusion are developed in a unified form of contrastive learning, where we pull semantically associated semantic units from different data signals together and push apart irrelevant ones in their representation spaces. 
Based on the pre-trained representations, we develop a recommender module and a conversation module to accomplish the corresponding tasks. Our approach is general to leverage various types of external data for CRS.

To our knowledge, it is the first time that multi-type external data has been leveraged for CRS. We address this task with a novel coarse-to-fine contrastive learning approach, which can better fuse semantic spaces from different data signals.
Such an approach can be extended to incorporate more kinds of external data. Extensive experiments on two public CRS datasets have demonstrated the effectiveness of our approach in both recommendation and conversation tasks.

\ignore{The major novelty lies in the coarse-to-fine pre-training stage. 
Specifically, we first devise a coarse-grained pre-training objective based on the macro user preference to unify and enhance the representations of multi-type input data. And then we devise a fine-grained pre-training objective based on the micro user preference to align and enhance the representations of the elements within the multi-type input data.
These optimization objectives are based in a unified form of contrastive learning.
In this way, we not only bridge the semantic gap among the different format data, but also capture and utilize the multi-grained preference within the multi-type data.
Based on the pre-trained representations, we develop a recommender module and a conversation module to accomplish the corresponding tasks.
Since our framework considers utilizing external structured and unstructured data, 
it is flexible to fuse various background information of users and items in online platform for improving the CRS task.
}

\ignore{The main contributions of this work are listed as follows:

(1) We propose a contrastive learning based pre-training framework to obtain effective representations for conversational recommendation task, which is flexible to fuse various background information of users and items in online platform for improving the CRS task.

2. Based on the multi-grained correlations within the input data, we devise a coarse-to-fine pre-training strategy to not only bridge the semantic gap among the different format data, but also capture and utilize the multi-grained preference within the multi-type data.

3. Extensive experiments on two public CRS datasets have demonstrated the effectiveness of our approach in both recommendation and conversation tasks.
}

%% file: related.tex
\section{Related Work}
In this section, we review the related work from the following two perspectives, namely conversational recommender system and contrastive learning.

\subsection{Conversational Recommender System}
With the rapid development of dialog system~\cite{Serban2015BuildingED,zhang2018personalizing,zhou2016multi,xu2019neural} in recent years, interactive conversation with users becomes an appealing approach to obtaining user dynamic intent and preference. 
Based on it, conversational recommender system~\cite{SunZ18,lei2020estimation,Li2018TowardsDC} (CRS) has become an emerging research topic, which aims to provide high-quality recommendations to users through natural language conversations.

One category of CRSs utilized predefined actions to interact with users, such as item attributes~\cite{Christakopoulou16,zhou2020leveraging} and intent slots~\cite{SunZ18}. Most of these methods mainly focus on accomplishing the recommendation task within fewer turns~\cite{SunZ18,lei2020estimation}. A surge of works adopt multi-armed bandit~\cite{Christakopoulou16}, reinforcement learning~\cite{lei2020estimation} and Thompson sampling~\cite{li2020seamlessly} for interaction with users.
However, this category of works do not pay much attention on generating high-quality natural language responses but usually rely on predefined dialog templates and rules to compose responses~\cite{SunZ18,lei2020estimation}.

Another category of CRSs develop natural language based approaches, focusing on both making accurate recommendation and generating human-like responses~\cite{Liao2019DeepCR,zhou2020improving,liang2021learning,wang2021finetuning}, these methods incorporate a generation-based dialog component to converse with users.
However, since a conversation usually contains a few sentences, lack of sufficient contextual information for accurately capturing user preference. Existing works leverage entity-oriented knowledge graph~\cite{Chen2019TowardsKR}, word-oriented knowledge graph~\cite{zhou2020improving} and review information~\cite{lu2021revcore} to alleviate this issue.

This paper extends the second category of research by leveraging rich external data for improving CRSs. 
Our key novelty lies in the coarse-to-fine contrastive learning approach, which can effectively fuse multi-type heterogeneous data in a principled way. 

\subsection{Contrastive Learning}
Contrastive learning has been long considered as effective in constructing meaningful representations~\cite{hadsell2006dimensionality,logeswaran2018efficient,chen2020simple}, which learns the representations by making a comparison between different samples.
Usually, it assumes a set of paired examples, where both examples are semantically related neighbors. 
Then the training objective is defined to pull the representations of neighbors together and push apart non-neighbors.

Recently, as a pre-training technique, contrastive learning has achieved remarkable success in computer vision~\cite{chen2020simple,he2020momentum}, natural language processing~\cite{wu2020clear,gao2021simcse} and information retrieval~\cite{bian2021contrastive}.
These works utilize data augmentation strategies to construct semantic related example pairs based on original data (\eg random cropping and rotation of images~\cite{chen2020simple}). 
Besides, contrastive learning has also been adopted to fuse multi-view information, such as image and text~\cite{zhang2021cross,nan2021interventional,zhang2020contrastive}, text and graph~\cite{huang2021learning}. It drives the representations of related information in different types to be similar, such that each kind of  representations can be enhanced by each other.

In our method, we first extract and represent associated multi-grained semantic units from different data signals, and then align the corresponding semantic units from different data signals via a coarse-to-fine contrastive learning.
Such a contrastive learning approach can effectively fuse the representations of these semantic units from heterogeneous signals. 

%% file: problem.tex
\section{Preliminaries}
The goal of CRS is to recommend appropriate items to a user via a multi-turn natural language conversation.
Typically, a CRS consists of two core components, namely the recommender component and the conversation component, and the two components should be integrated seamlessly to fulfill the recommendation goal.  

\paratitle{Notations for CRS.} Formally, let $u$ denote a user from user set $\mathcal{U}$, $i$ denote an item from item set $\mathcal{I}$, and $w$ denote a word from vocabulary $\mathcal{V}$.
A conversation (or a conversation history) $C$ consists of a list of utterances, denoted by $C=\{s_{t}\}_{t=1}^{n}$, in which each utterance $s_t$ is a conversation sentence at the $t$-th turn and each utterance ${s_{t}}=\{w_{j}\}_{j=1}^{m}$ is composed by a sequence of words.
As the conversation goes on, the conversation utterances are aggregated (called \emph{conversation history} or \emph{conversation context}), and the CRS estimates the user preference based on the conversation history. 
At the $t$-th turn, the recommender component selects a set of candidate items $\mathcal{I}_{t+1}$ from the entire item set $\mathcal{I}$ according to the estimated user preference, while the dialog component needs to produce the next utterance $s_{t+1}$ to reply to previous utterances.

\paratitle{External Data.} Besides the conversation history $C$, in the online platform, some external data is usually available to enhance the performance of CRS.
In the literature, two kinds of external data are widely used for CRS, namely structured external data \emph{knowledge graph}~\cite{Chen2019TowardsKR,zhou2020improving} and unstructured external data \emph{item reviews}~\cite{lu2021revcore}.
We also consider leveraging the two kinds of external data in this work. 
For knowledge graph $\mathcal{G}$, it is composed by an entity set $\mathcal{N}$ and a relation set $\mathcal{R}$. The entity set contains all the items (\ie $\mathcal{I}$ is a subset of the entity set) and other item-related entities.
Furthermore, for each user $u$, her/his interacted entities set (\ie entities occurring in the conversation history) is denoted by $\mathcal{N}_{u}$, which composes a subgraph $\mathcal{G}_{u}$ that reflect the user preference.
For an online review, it can be considered as a document consisting of sentences $\mathcal{S}=\{\tilde{s}_{j}\}_{j=1}^{n}$.
The two kinds of external data covers both structured and unstructured data, and reflect a multi-grained semantic form: entity \emph{v.s.} subgraph for knowledge graph, and sentence \emph{v.s.} document for online reviews. 
As will be shown later, our approach is general to incorporate multiple types of external data in CRS.

\paratitle{Task Definition.} Based on these notations, the task of this paper is defined as: given the multi-type context data (\ie conversation history $C$, knowledge graph $\mathcal{G}$ and reviews $\mathcal{S}$), we aim to (1) accurately recommend items $\mathcal{I}_{t+1}$ and (2) generate proper response $s_{t+1}$ to the user $u$ at the $(t+1)$-th turn of a conversation.

%% file: model.tex
\section{Approach}
\label{CRS}

\begin{figure*}
\includegraphics[width=1.0\textwidth]{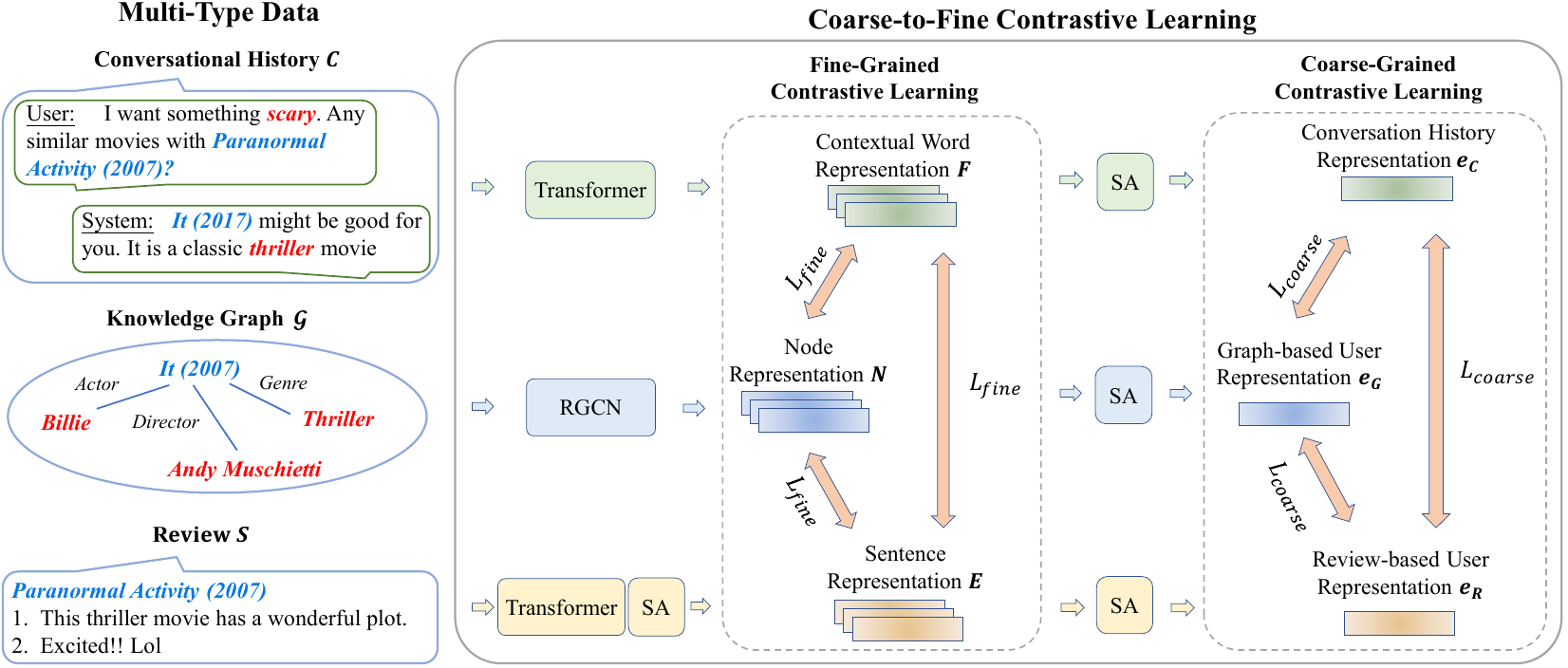}
\caption{The overview of our model in a movie recommendation scenario, where ``SA'' denotes the self-attention layer. In the coarse-to-fine contrastive learning, we first conduct coarse-grained contrastive learning using coarse-grained features, and then conduct fine-grained contrastive learning using fine-grained features.}
\label{approach}
\end{figure*}

In this section, we present the proposed \underline{\textbf{C}}oarse-to-fine \underline{\textbf{C}}ontrastive learning for \underline{\textbf{C}}onversational recommender system, called \textbf{C$^2$-CRS}. 
Since we consider utilizing multi-type context data for CRS, we first study how to separately encode these context data. 
Then, we introduce the coarse-to-fine contrastive learning approach to gradually fusing context data for pre-training effective data representations. 
Based on the learned representations, we finally describe our solutions for both recommendation and conversation tasks.
The overview illustration of the proposed model is presented in Figure~\ref{approach}.

\subsection{Encoding Multi-type Context Data}
\label{Sec-encoder}
Given the multi-type context data, we adopt corresponding encoding modules to generate the data representations in separate semantic spaces. Next, we present the encoding modules for conversation history, knowledge graph and reviews, respectively.

\subsubsection{Encoding Conversation History} 
\label{Sec-encodeC}
Conversation history $C$ consists of utterances $\{s_{t}\}_{t=1}^{n}$ generated in a session between the CRS and the user, which depicts the user-system interaction about the information needs.
Since the utterances, which are usually short sentences, are closely relevant, following previous works~\cite{Chen2019TowardsKR,zhou2020improving}, we concatenate these utterances $\{s_{t}\}_{t=1}^{n}$ in a chronological order to form a long sentence $s_{1:n}$.
To obtain the representations of the conversation history, we adopt a standard Transformer~\cite{vaswani2017attention} to encode $s_{1:n}$ as following:
\begin{align}
    \bm{F} &= \text{Transformer}(s_{1:n}), \label{eq:Fl}
\end{align}
where contextual word representations $\bm{F}$ are obtained on the top layer of Transformer.

Then we utilize a self-attentive layer to produce the representation of the conversation history $\be_{C}$:
\begin{align}\label{eq-SA}
\bm{\bm{e}}_{C} &= \bm{F} \cdot \text{softmax}( \bm{b}^\top \text{tanh}(\mathbf{W}_{sa}  \bm{F})),
\end{align}
where $\mathbf{W}_{sa} $ and $\bm{b}$ are learnable parameter matrix and vector.

\subsubsection{Encoding Knowledge Graph}
The knowledge graph $\mathcal{G}$ consists of an entity set $\mathcal{N}$ and a relation set $\mathcal{R}$. It stores semantic facts in the form of a triple $\langle n_1,r,n_2\rangle$.

Consider that the edges connected the nodes may own useful relation information, we utilize R-GCN~\cite{Schlichtkrull_2018} to encode $\mathcal{G}$. 
Formally, the representation of node $e$ at $(l+1)$-th layer is calculated as:
\begin{equation}
\label{r-gcn}
\bm{n}_{e}^{(l+1)}=\sigma\big(\sum_{r \in \mathcal{R}}\sum_{e' \in \mathcal{E}_{e}^{r}}\frac{1}{Z_{e,r}}\mathbf{W}_{r}^{(l)}\bm{n}_{e'}^{(l)}+\mathbf{W}^{(l)}\bm{n}_{e}^{(l)}\big)
\end{equation}
where $\bm{n}_{e}^{(l)} \in \mathbb{R}^{d_E}$ is the node representations of $e$ at the $l$-th layer, $\mathcal{E}_{e}^{r}$ denotes the set of neighboring nodes for $e$ under the relation $r$, $\mathbf{W}_{r}^{(l)}$ and $\mathbf{W}^{(l)}$ are learnable matrices and $Z_{e,r}$ is a normalization factor.
After aggregating the graph information, we obtain the node representations matrix $\bm{N}$ on the top R-GCN layer.
As shown in previous studies~\cite{Chen2019TowardsKR,zhou2020improving}, it is particularly important to consider the historically interacted entities $\mathcal{N}_{u}$ for modeling the user preference. Therefore, we collect the node representations of user interacted entities as $\bm{N}_{u}$, and utilize self-attentive mechanism as Eq.~\ref{eq-SA} to generate the graph-based user representation ${\bm{e}}_{G}$. The basic idea is to automatically learn the importance of each interacted entity for user $u$, so that we can derive the user preference considering the levels of entity importance.

\subsubsection{Encoding Reviews}
The review text $\mathcal{S}_{i}$ of an item is a set of sentences $\mathcal{S}_{i}=\{\tilde{s}_{j}\}_{j=1}^{n}$ written by online users about the item $i$. 
Similar as the conversation history, we utilize the standard Transformer model to encode each sentence and leverage a self-attention layer to obtain the sentence representation matrix $\bm{E}$.
Note that the parameters of this Transformer and self-attention layer are different from those in Section~\ref{Sec-encodeC}.
Besides, consider that the reviews are usually noisy and may be irrelevant to user preference~\cite{lu2021revcore}, we further utilize a sentence-level self-attention layer to select the useful information within the sentence representations.
Finally, we produce the review-based user representation $\bm{e}_{R}$.

After the above encoding, we can obtain the corresponding representations for conversation history, knowledge graph and review text. The three kinds of context data are represented in different semantic spaces. Next, we study how to fuse them in order to derive shared data semantics.

\subsection{Coarse-to-Fine Contrastive Learning}
In order to utilize context data, we need to effectively fuse their representation spaces, so that their information can be shared and leveraged across different data signals. 
In this section, based on the multi-grained correlations between multi-type context data, we propose a coarse-to-fine contrastive learning method to fuse the multi-type information to enhance the data representations.
The main idea is to represent each type of data from both coarse-grained and fine-grained views and associate the corresponding representations at different granularities in a coarse-to-fine manner. 
We will introduce the coarse-grained and fine-grained pre-training approaches in the following.

\subsubsection{Coarse-Grained Contrastive Learning} 
\label{coarse-grained pre}
Multi-type context data in a coarse-grained form mainly reflects the overall user preference. 
As in Section~\ref{Sec-encoder}, we obtain the coarse-grained representations of the user from the views of the conversation data $\bm{e}_{C}$ (\ie conversation history), knowledge graph $\bm{e}_{G}$ (\ie subgraph) and review text $\bm{e}_{R}$ (\ie review document) .
There exists a natural semantic gap between these data representations. 
To bridge this gap, we propose a contrastive learning based pre-training approach to aligning and fusing the three types of context data.

Contrastive learning is a widely adopted pre-training technique which learns the representations by pulling semantically close representations together and pushing apart non-related ones. 
In our setting, we have three representations $\bm{e}_{C}$, $\bm{e}_{G}$ and $\bm{e}_{R}$ depicting the same user preference from different views, which are considered as semantically close ones. 
Therefore, based on the three representations, we take the pairs  $(\bm{e}_{C},\bm{e}_{G})$, $(\bm{e}_{C},\bm{e}_{R})$, and $(\bm{e}_{G},\bm{e}_{R})$ as positive examples, while the representations from different users in the same batch are considered as negative examples.
Thus, for a mini-batch with $b$ pairs, the coarse-grained pre-training objective is the sum of the triple contrast learning loss, which can be formulated as:
\begin{eqnarray}
    {L}_{coarse} &=& L_{CL}(\bm{e}_{C},\bm{e}_{G}) + L_{CL}(\bm{e}_{C},\bm{e}_{R}) + L_{CL}(\bm{e}_{G},\bm{e}_{R}),\\
    L_{CL}(\bm{h}, \bm{h}^+) &=& \log\frac{e^{\text{sim}(\bm{h},\bm{h}^{+})/\tau}}{\sum_{\bm{h}^{-}_i\in\left\{\bm{h}^-\right\}}{e^{\text{sim}(\bm{h},\bm{h}^{-}_i)/\tau}}},
\label{eq-cl}
\end{eqnarray}
where $\bm{h}$ and $\bm{h}^{+}$ are two types of coarse-grained representations of a user, $\left\{\bm{h}^-\right\}$ is the negative example set for positive examples $(\bm{h}, \bm{h}^+)$, $\tau$ is a temperature hyperparameter and $\text{sim}(\bm{h}, \bm{h}^+)$ is the cosine similarity $\frac{\bm{h}^{\top}h^+}{||\bm{h}||\cdot||\bm{h}^{+}||}$. 
Since the representations of the same users under different views are pulled together, this objective aligns the semantic space of the three types of context data. 
The three types of representations capture user preference from different views, which are complementary to each other.
By semantic alignment and fusion using the contrastive learning objective, these representations can be also mutually improved by each other.

\subsubsection{Fine-Grained Contrastive Learning}
The above coarse-grained contrastive learning fuses the semantic space at the overall level. However, the corresponding semantic associations between fine-grained semantic units (\eg words and entities) are neglected.
Fine-grained context data captures specific user preference about fine-grained characteristics. 
Therefore, we further propose to conduct fine-grained contrastive learning for better fusing the representation spaces. 
For the three types of context data, the fine-grained preferences are encoded in the contextual word representation (${\bm{F}}_{j}$) of a word from conversation history, the node representation (${\bm{N}}_{e}$) of an entity from the knowledge graph, and the representation (${\bm{E}}_{k}$) of a sentence from the review document, respectively.

Similar to the coarse-grained pre-training, we also adopt the contrastive learning method for fine-grained pre-training.
To construct the semantic-closed example pairs, we consider capturing the correlations among these fine-grained semantic units, \ie words from conversation history, entities from knowledge graph and sentences from review text.
We also consider inter-type data associations. 
For the word-entity correlation, we utilize the entity linking method~\cite{Chen2019TowardsKR} to match the entity from the knowledge graph with its corresponding word in the conversation history.
For the entity-sentence correlation, we match the entity with its most related sentence, which may be the most highly-rate review or descriptive text of the entity.
In this way, we can construct the semantically-consistent \emph{word-entity-sentence} representation triples $({\bm{F}_{j}},{\bm{N}}_{e},{\bm{E}}_{k})$ to depict the same fine-grained user preference.
Then, we split the triples into pairs $({\bm{F}_{j}}, {\bm{N}}_{e})$, $({\bm{F}_{j}}, {\bm{E}}_{k})$ and $({\bm{N}}_{e}, {\bm{E}}_{k})$ to construct paired positive examples for contrastive learning.
Similarly, the representations from other users in the same batch compose a negative example set. 

Therefore, our fine-grained pre-training objective can be formulated as:
\begin{align}
    {L}_{fine} = L_{CL}({\bm{F}_{j}}, {\bm{N}}_{e}) + L_{CL}({\bm{F}_{j}}, {\bm{E}}_{k}) + L_{CL}({\bm{N}}_{e}, {\bm{E}}_{k}),
\end{align}
where $L_{CL}(\cdot)$ is defined according to Eq.~\ref{eq-cl}.
This optimization objective enforces the alignment among semantic spaces of words, nodes and sentences.

To further preserve the enhancement effect of coarse-grained contrastive learning, we integrate its optimization objective with a weight   $\lambda$. 
Then, the final fine-grained pre-training objective ${L}$ is given as follows:
\begin{align}
    \label{fine-pre-loss}
    {L} = {L}_{fine} + \lambda \cdot {L}_{coarse}.
\end{align}
Instead of directly fusing different semantic spaces, we design a coarse-to-fine fusion way, so the semantic space will be gradually pulled close and finally fused.  

\subsection{Fine-tuning the Conversation Recommender System}
Based on the pre-trained representations, we introduce our approach to fine-tuning the data representations and network architectures to accomplish the conversational recommendation task.
In the following, we will describe the architectures of our recommendation module and response generation module, and detail how to fine-tune them.

\subsubsection{Fine-tuning Recommendation Module} 
Given the pre-trained item representations, we study how to fine-tune them for the recommendation task. 
We first generate the user representation, and then fine-tune it on item recommendation task.

Note that after semantic fusion, the learned entity representations have learned useful information from other types of context data. 
Therefore, we only adopt the entities $\mathcal{N}_{u}$ occurring in the conversation history for learning the user representation.
Specifically, we stack the entities' representations of $\mathcal{N}_{u}$ into a matrix $\bm{N}_{u}$, and then utilize self-attentive mechanism as in Eq.~\ref{eq-SA} to produce the user representation $\bm{e}_{u}$.
\ignore{
Following previous works~\cite{Chen2019TowardsKR,zhou2020improving}, we mainly utilize the node representations $\bm{N}_{u}$ (Eq.~\ref{r-gcn}) from the interacted entities  to produce the user representation.
Note that after semantic fusion, the learned entity representations are injected into useful information from other types of context data. 
To accurately capture the user current preference, we only adopt the entities $\mathcal{N}_{u}$ occurring in the conversation history for learning the user representation.
Specifically, we stack the entities' representations of $\mathcal{N}_{u}$ into a matrix, and then utilize self-attentive mechanism as in Eq.~\ref{eq-SA} to produce the user representation $\bm{e}_{u}$.}
Finally, we compute the probability that recommends an item $i$ from the item set to a user $u$:
\begin{equation}
\label{eq-rec}
    \text{P}_{rec}(i) = \text{softmax}(\bm{e}_{u}^\top \bm{n}_{i}),
\end{equation}
where $\bm{n}_i$ is the learned item embedding for item $i$. We can utilize Eq.~\ref{eq-rec} to rank all the items and generate a recommendation set to a user.
To fine-tune the user representation $\bm{e}_{u}$ and item embedding $\bm{n}_i$, we apply a cross-entropy loss as the optimization objective:
\begin{eqnarray}
\label{CE-rec}
    L_{rec}=-\sum_{j=1}^{N}\sum_{i=1}^{M} y_{ij}\cdot\log\big(\text{P}^{(j)}_{rec}(i)\big),
\end{eqnarray}
where $N$ is the number of conversations, $j$ is the index of a conversation, $M$ is the number of items, and $i$ is the index of an item. Here, we consider the case with a single ground-truth recommendation. It will be easy to extend the above loss to the case with multiple ground-truth items.

\subsubsection{Fine-tuning Response Generation Module} 
Here, we study how to fine-tune the pre-trained representations for the conversation task.
Following KGSF~\cite{zhou2020improving}, we incorporate multiple cross-attention layers in a standard Transformer~\cite{vaswani2017attention} decoder to fuse the pre-trained representations as following:
\begin{eqnarray}
\mathbf{R}^{l}=\text{Decoder}(\mathbf{R}^{l-1}, \bm{N}, \bm{E}, \bm{F}),\label{eq-T}
\end{eqnarray}
where $\mathbf{R}^{l}$ is the output representation matrix from the decoder at $l$-th layer.
The above equation follows the similar transformation chain as KGSF~\cite{zhou2020improving}:
\emph{generated words} $\rightarrow$ \emph{conversation history} $\rightarrow$ \emph{knowledge graph} $\rightarrow$ \emph{reviews}.
We omit the equation details.
Based on it, following existing works~\cite{zhou2020improving}, we also design the copy network to enhance the generation of informative response.
\ignore{
Formally, given the predicted subsequence $y_1, \cdots, y_{i-1}$, the probability of generating $y_i$ as the next token is given as:
\begin{eqnarray}
\text{Pr}(y_i | y_1, \cdots, y_{i-1} ) = \text{Pr}_{1}(y_i | \mathbf{R^{'}_{i}})
\label{gen-copy}
\end{eqnarray}
where $\text{Pr}_{1}(\cdot)$ is the generative probability implemented as a softmax function over the vocabulary by taking the decoder output $\mathbf{R^{'}}_i$ (Eq.~\ref{enhanced_decoder_out}) enhanced by external unstructured data as input.
The enhanced decoder output $\mathbf{R^{'}}_i$ can be obtained by:
\begin{align}\label{enhanced_decoder_out}
\bm{\alpha} &= \text{softmax}( \mathbf{R^{'}}_i \cdot \mathbf{{E}_{un}}^\top ),\\
\mathbf{\tilde{E}_{un}} &= \mathbf{{E}_{un}} \cdot \bm{\alpha},\\\nonumber
\mathbf{R^{'}} &= \left[ \mathbf{R^{'}}, \mathbf{\tilde{E}_{un}} \right] \bW + \bb
\end{align}
where $\bW$, $\bb$ are trainable parameters, $\bm{E}_{Un}$ is fine-grained external unstructured sentence representations, $\bm{\alpha}$ is an attention weight vector reflecting the importance of each external unstructured sentence.}

To fine-tune this module for generating more informative responses, we devise an instance weighting enhanced cross-entropy loss as:
\begin{eqnarray}
L_{gen} &=& -\frac{1}{m}\sum_{j=1}^{m}\log\big({\alpha_{w_{j}}} \text{P}(w_{j} |w_1, \cdots, w_{j-1}))\big), \\
\alpha_{w_{j}} &=& \begin{cases} \text{max}(\gamma, \frac{\beta}{f_{w_{j}}}) \quad \text{if} \ f_{w_j} \geq \beta \\ 1 \quad \  \quad \quad \quad \quad \text{otherwise} \end{cases} ,
\label{CE-gene}
\end{eqnarray}
where $m$ is the number of words in generated response, ${{\alpha}_{w_{j}}}$ is the weight considering the frequency of this token, $\beta$ is a preset threshold, $\gamma$ is to avoid punishing the high-frequency words too much, and $f_{w_{j}}$ is the word frequency of ${w_{j}}$ in corpus.
With the above cross-entropy loss, we can punish the high-frequency tokens, and help generate more informative responses.

\subsection{Discussion}
To summarize, our approach provides a novel coarse-to-fine contrastive learning framework for leveraging multi-type context data to improve CRS. 
Next, we compare it with existing studies. 

\textbf{Conversation-centered approaches} 
such as ReDial~\cite{Li2018TowardsDC} and CRM~\cite{SunZ18} mainly focus on using the conversation history to accomplish the conversational recommendation task, with or without the external auxiliary data.
Our work falls into this category and extends it by leveraging multi-type context data with a novel coarse-to-fine contrastive learning framework. As a comparison, most of existing studies focus on some specific type of external data, which is not general to leverage various heterogeneous external data.  

\textbf{Knowledge graph based approaches} 
such as DeepCR ~\cite{Liao2019DeepCR}, KBRD~\cite{Chen2019TowardsKR} and KGSF ~\cite{zhou2020improving} fuse external KG and conversational context to help model user representations and generate more informative responses. This category of studies only considers structured external data, which neglects rich unstructured data as below.

\textbf{Review-enhanced approaches}
such as RevCore~\cite{lu2021revcore} introduce external reviews to CRS as the supplement of conversational context. These methods extract entities from reviews for recommender module and utilize copy mechanism for conversation module.
However, reviews are inevitable to contain noise.
In our approach, we leverage the coarse-to-fine pre-training approach to fuse the multi-type data, which can adaptively fuse useful information from data and prevent the noisy reviews to directly affect the modeling of user preference.

%% file: experiment.tex
\section{Experiment}
In this section, we first set up the experiments, and then report the results and give a detailed analysis. 

\subsection{Experiment Setup}
In this subsection, we provide an introduction to the details of our experiments, including dataset, baselines, evaluation metrics and implementation details.
\begin{table}
\caption{Results on the recommendation task. Numbers marked with * indicate that the improvement is statistically significant compared with the best baseline (t-test with p-value $< 0.05$).}\label{rec-result}
\small
\centering
\begin{tabular}{l|ccc|ccc}
    Dataset & \multicolumn{3}{c}{ReDial}  & \multicolumn{3}{c}{TG-ReDial}\\
    \hline
     Models &R@1 &R@10 &R@50 &R@1 &R@10 &R@50\\
    \hline
    Popularity & 0.011 & 0.054 & 0.183 & 0.0004 & 0.003 & 0.014\\
    TextCNN & 0.013 & 0.068 & 0.191 & 0.003 & 0.010 & 0.024 \\
    ReDial & 0.024 & 0.140 & 0.320 &  0.000 & 0.002 & 0.013 \\
    KBRD & 0.031 & 0.150 & 0.336 & 0.005 & 0.032 & 0.077 \\
    KGSF & 0.039 & 0.183 & 0.378  & 0.005 & 0.030 & 0.074\\
    KECRS & 0.021 & 0.143 & 0.340 & 0.002 & 0.026 & 0.069 \\
    RevCore & 0.046 & 0.220 & 0.396 & 0.004 & 0.029 & 0.075 \\
    \hline
    \textbf{C$^2$-CRS} & \textbf{0.053}*& \textbf{0.233}*& \textbf{0.407}* & \textbf{0.007}*& \textbf{0.032}*& \textbf{0.078}* \\
    \hline
  \end{tabular}
\end{table}

\subsubsection{Dataset}
We evaluate our model on \textsc{ReDial}~\cite{Li2018TowardsDC} and \textsc{TG-ReDial}~\cite{zhou2020towards} datasets. 
The \textsc{ReDial} dataset is an English conversational recommendation dataset constructed with Amazon Mechanical Turk (AMT). Following a set of comprehensive instructions, the AMT workers played the roles of seekers and recommenders to generate dialogue for recommendation on movies. It contains 10,006 conversations consisting of 182,150 utterances related to 51,699 movies.
The \textsc{TG-ReDial} dataset is a Chinese conversational recommendation dataset, which emphasizes natural topic transitions from non-recommendation scenarios to the desired recommendation scenario. It is created in a semi-automatic way, hence human annotation is more reasonable and controllable. It contains 10,000 two-party dialogues consisting of 129,392 utterances related to 33,834 movies. 
For each conversation, we start from the first sentence one by one to generate reply to utterances or give recommendations. Since the above datasets do not contain the review data, we retrieve reviews for movies in \textsc{ReDial} and \textsc{TG-ReDial} from IMDB~\footnote{https://www.dbpedia.org/} and douban~\footnote{https://movie.douban.com/} respectively. 

\subsubsection{Baselines}
In CRS, we consider two major tasks to evaluate our method, namely recommendation and conversation. Therefore, we not only compare our approach with existing CRS methods, but also select representative recommendation or conversation models as baselines. 

$\bullet$ \emph{Popularity}: It ranks the items according to recommendation frequencies in the training set of the corpus.

$\bullet$ \emph{TextCNN}~\cite{Kim14}: It applies a CNN-based model to extract user features from conversational context for ranking items.

$\bullet$ \emph{Transformer}~\cite{vaswani2017attention}: It utilizes a Transformer-based encoder-decoder method to generate conversational responses.

$\bullet$ \emph{ReDial}~\cite{Li2018TowardsDC}: This model is proposed in the same paper with the \textsc{ReDial} dataset. It consists of a dialog generation module based on HRED~\cite{serban2017hierarchical} and a recommender module based on auto-encoder~\cite{he2017distributed}.

$\bullet$  \emph{KBRD}~\cite{Chen2019TowardsKR}: This model utilizes DBpedia to enhance the semantics of contextual items. The Transformer architecture is applied in the dialog generation module, in which KG information is used as word bias in generation.

$\bullet$  \emph{KGSF}~\cite{zhou2020improving}: This model incorporates DBpedia and ConceptNet to enhance the semantic representations of items and words, and uses Mutual Information Maximization to align the semantic spaces of different components. 

$\bullet$  \emph{KECRS}~\cite{zhang2021kecrs}: It constructs a high-quality KG and develops the Bag-of-Entity loss and the infusion loss to better integrate KG with CRS for generation.

$\bullet$  \emph{RevCore}~\cite{lu2021revcore}: It proposes a review-enhanced framework, in which reviews are selected by a sentiment-aware retrieval module and are utilized to enhance recommender module and dialogue generation module. For a fair comparison, we use the same review as our approach.

Among these baselines, \emph{Popularity} and \emph{TextCNN}~\cite{Kim14} are recommendation methods, while \emph{Transformer}~\cite{vaswani2017attention} is the state-of-the-art text generation method. 
We do not include other recommendation models since there are no historical user-item interaction records except dialogue utterances.
Besides, \emph{REDIAL}~\cite{Li2018TowardsDC}, \emph{KBRD}~\cite{Chen2019TowardsKR}, \emph{KGSF}~\cite{zhou2020improving}, \emph{KECRS}~\cite{zhang2021kecrs} and \emph{RevCore}~\cite{lu2021revcore} are conversational recommendation methods. 
We name our proposed model as \emph{C$^2$-CRS}.

\subsubsection{Evaluation Metrics}
In our experiments, we adopt different metrics to evaluate on the two tasks.
For the recommendation task, following~\cite{Chen2019TowardsKR}, we adopt Recall@$k$ ($k=1, 10, 50$) for evaluation.
For the conversation task, the evaluation consists of automatic evaluation and human evaluation. 
Following~\cite{Chen2019TowardsKR}, we use Distinct $n$-gram ($n=2,3,4$) to measure the diversity at sentence level. 
Besides, we invite three annotators to score the generated candidates in two aspects, namely \emph{Fluency} and \emph{Informativeness}. The range of score is 0 to 2. The final performance is calculated using the average score of three annotators.

\subsubsection{Implementation Details} 
We implement our approach with Pytorch~\footnote{https://pytorch.org/} and CRSLab~\cite{zhou-etal-2021-crslab}~\footnote{https://github.com/RUCAIBox/CRSLab}. 
The dimensionality of embeddings (including hidden vectors) is set to 300 and 128, respectively, for conversation and recommender modules. In the structured encoder module, we set the layer number to 1 for GNN networks.
We use Adam optimizer~\cite{KingmaB14} with the default parameter setting. In experiments, the batch size is set to 256, the learning rate is 0.001, gradient clipping restricts the gradients within $[0, 0.1]$, the temperature of contrastive learning is set to 0.07, and the normalization constant $Z_{v,r}$ of R-GCN in Eq.~\ref{r-gcn} is 1. During the coarse-grained pre-training stage, we directly optimize the loss as Section~\ref{coarse-grained pre}. While in the fine-grained pre-training stage, the weight $\lambda$ of the coarse-grained contrastive learning loss in Eq.~\ref{fine-pre-loss} is set to 0.2. The code and data are available at: \textcolor{blue}{\url{https://github.com/RUCAIBox/WSDM2022-C2CRS}}.

\subsection{Evaluation on Recommendation Task}
To verify the effectiveness of our proposed method on the recommendation task, we conduct a series of experiments and present the results in Table~\ref{rec-result}.

In general, conversational recommendation methods perform better than recommendation methods (\eg TextCNN and Popularity). Since these methods mainly focus on integrating the conversational module and the recommendation module, which are mutually beneficial for each other.
For recommendation methods, we can see that TextCNN achieves better performance than Popularity. One reason is that Popularity only recommends the most popular items without considering the contextual information. In contrast, TextCNN is able to model personalized preference from contextual text for better recommendation.

For conversational recommendation methods, first, compared with the \textsc{TG-ReDial} dataset, ReDial model performs better on the \textsc{ReDial} dataset. One possible reason is that the contextual items in conversation from the \textsc{TG-ReDial} dataset are much sparser than those in \textsc{ReDial} dataset, but ReDial model relies heavily on the contextual items to generate the recommendation.
Second, KBRD achieves better performance than ReDial. Since KBRD utilizes knowledge graph as external contextual information to improve user preference modeling, and then utilizes R-GCN and Transformer to accomplish the recommendation and conversation tasks, respectively.
KGSF achieves better performance than KBRD and KECRS. It improves the data representations by aligning the two semantic spaces between conversations and items via mutual information maximization.
Finally, RevCore achieves better performance than other baselines. It incorporates external reviews to enhance the description of items, which help better capture user preference.

Our model C$^2$-CRS outperforms all the baselines, since C$^2$-CRS utilizes multi-type external information to help understand the conversational history, including conversational text, knowledge graph and reviews.
To achieve it, our approach applies the coarse-to-fine contrastive learning to gradually fuse different types of information. Such a way can be beneficial to the recommender module by enhancing the data representations.

\begin{table}
\caption{Automatic evaluation results on the conversation task, where "Transf" denotes Transformer model. We abbreviate Distinct-2,3,4 as Dist-2,3,4. Numbers marked with * indicate that the improvement is statistically significant compared with the best baseline (t-test with p-value $< 0.05$).}\label{gen-result}
\centering
\begin{tabular}{l|ccc|ccc}
     Dataset & \multicolumn{3}{c}{ReDial} & \multicolumn{3}{c}{TG-ReDial} \\
    \hline
     Models &Dist-2 &Dist-3 &Dist-4 &Dist-2 &Dist-3 &Dist-4\\
    \hline
    Transf & 0.067 & 0.139 & 0.227 & 0.053 & 0.121 & 0.204 \\
    ReDial & 0.082 & 0.143 & 0.245 & 0.055 & 0.123 & 0.215 \\
    KBRD & 0.086 & 0.153 & 0.265 & 0.045 & 0.096 & 0.233  \\
    KGSF & 0.114 & 0.204 & 0.282 & 0.086 & 0.186 & 0.297 \\
    KECRS & 0.040 & 0.090 & 0.149 & 0.047 & 0.114 & 0.193 \\
    RevCore & 0.092 & 0.163 & 0.221 & 0.043 & 0.105 & 0.175\\
    \hline
    \textbf{C$^2$-CRS} & \textbf{0.163}*& \textbf{0.291}*& \textbf{0.417}* & \textbf{0.189}*& \textbf{0.334}*& \textbf{0.424}* \\
    \hline
  \end{tabular}
\end{table}

\subsection{Evaluation on Conversation Task}
In this subsection, we verify the effectiveness of the proposed model for the conversation task and report the results on automatic and human evaluation metrics.

\subsubsection{Automatic Evaluation}
We present the results of the automatic evaluation metrics for different methods in Table~\ref{gen-result}.

First, we can see that ReDial performs better than Transformer, since ReDial applies a pre-trained RNN model to produce better representations of historical conversation.
Second, KBRD achieves better performance than ReDial in most settings. Since it enhances contextual entities and items by external KG and these entities are utilized to produce word probability bias for conversational module.
Third, KGSF generates the most diverse response among these baselines. Since it not only aligns the  1 of conversational text and items, but also enhances their representations. Besides, multiple cross-attention layers are performed in the Transformer decoder to further interact the contextual information with the generated response.
Finally, RevCore performs not well. One possible reason is that its involved reviews may contain noise.

Compared with these baselines, our model C$^2$-CRS performs consistently better in all the evaluation metrics, and it improves the Transformer decoder with enhanced multi-type data representations by coarse-to-fine contrastive learning. 
Such a gradual fusion approach is robust to the noise in the contextual information and can better capture useful semantics from conversational text, knowledge graphs, and reviews.
Besides, we further design an instance weighting mechanism to help generate more informative responses. It shows that our approach can effectively improve the response generation task.

\begin{table}
\caption{Human evaluation results on the conversation task, where Transf refers to the Transformer model. Numbers marked with * indicate that the improvement is statistically significant compared with the best baseline (t-test with p-value $< 0.05$). }\label{gen-result-human}
\centering
\begin{tabular}{l|cc}
     Models &Fluency &Informativeness\\
    \hline
    Transf &0.97 & 0.92 \\
    ReDial &1.35 & 1.04 \\
    KBRD  &1.23 & 1.15 \\
    KGSF & 1.48 &  1.37   \\
    KECRS & 1.39 & 1.19   \\
    RevCore & 1.52 & 1.34   \\
    \hline
    \textbf{C$^2$-CRS} & \textbf{1.55}*& \textbf{1.47}* \\
    \hline
  \end{tabular}
\end{table}

\subsubsection{Human Evaluation}
Table~\ref{gen-result-human} presents the results of the human evaluation for the conversation task.

First, ReDial achieves better performance than Transformer, since it incorporates a pre-trained RNN encoder~\cite{DBLP:conf/iclr/SubramanianTBP18}. 
Second, KBRD achieves a comparable performance with ReDial, which indicates that the entity information from KG is beneficial to informativeness. 
Third, among these baselines, KGSF performs the best in term of \emph{informativeness} since it aligns the conversational text and items via knowledge graph semantic fusion.
Fourth, RevCore performs the best in term of \emph{fluency}, since it utilizes external reviews to enhance the decoder for generating more fluent responses.

Finally, C$^2$-CRS performs the best in both metrics. By effectively leveraging and fusing multi-type data, our model is able to generate more informative words or entities, and meanwhile maintains the fluency of the generated text.
Besides, the instance weighting mechanism in the fine-tuning stage also helps generate more fluent and informative tokens.

\subsection{Ablation Study}
\begin{table}
\caption{Results of ablation and variation study on the recommendation task. Coarse and Fine refer to coarse-grained and fine-grained contrastive learning, respectively. CH, SD and UD refer to conversational history, external structured data and external unstructured data, respectively.}\label{ablation}
\centering
\begin{tabular}{l|ccc}
    Dataset & \multicolumn{3}{c}{ReDial} \\
    \hline
     Models &R@1 &R@10 &R@50\\
    \hline
    \textbf{C$^2$-CRS} & \textbf{0.053}*& \textbf{0.233}*& \textbf{0.407}* \\
    \hline
    C$^2$-CRS w/o Coarse-Fine & 0.031 & 0.150 & 0.336 \\
    C$^2$-CRS w/o Coarse & 0.049 & 0.220 & 0.394  \\
    C$^2$-CRS w/o Fine & 0.048 & 0.221 & 0.400  \\
    C$^2$-CRS Multi-task & 0.052 & 0.221 & 0.404 \\
    C$^2$-CRS w/o CH & 0.048 & 0.223 & 0.403 \\
    C$^2$-CRS w/o SD & 0.051 & 0.226 & 0.403  \\
    C$^2$-CRS w/o UD & 0.051 & 0.222 & 0.395 \\
    \hline
  \end{tabular}
\end{table}
We also conduct the ablation study based on different variants of our model, including: (1) \emph{C$^2$-CRS w/o Coarse-Fine} removes the coarse-to-fine contrastive learning; (2) \emph{C$^2$-CRS w/o Coarse} removes the coarse-grained contrastive learning; (3) \emph{C$^2$-CRS w/o Fine} removes the fine-grained contrastive learning; (4) \emph{C$^2$-CRS Multi-task} combines all the pre-training and fine-tuning tasks as a multi-task training; (5) \emph{C$^2$-CRS w/o CH} removes the conversational history; (6) \emph{C$^2$-CRS w/o SD} removes the structured data (\ie knowledge graph); and (7) \emph{C$^2$-CRS w/o UD} removes the unstructured data (\ie reviews).

As shown in Table~\ref{ablation}, firstly, we can see that removing coarse-to-fine contrastive learning leads to the largest performance decrease. Since there is a natural semantic gap in the multi-type data,
it is key to fuse the underlying semantics for effective information utilization.
Secondly, 
the direct multi-task setting (combining all the pre-training and fine-tuning tasks) yields a worse performance than our approach. The major reason is that it is difficult to fuse different types of data, which requires a more principled approach for semantic fusion.  
Finally, the variants that remove any kind of external data lead to a performance decrease. It shows that all kinds of external data are useful in our approach in enhancing the data representations.